\documentclass{article}

\usepackage{microtype}
\usepackage{graphicx}
\usepackage{subcaption}
\usepackage{booktabs} 

\usepackage{hyperref}

\newcommand{\githublogo}[1]{\hspace{2pt}\includegraphics[width=13pt, valign=c]{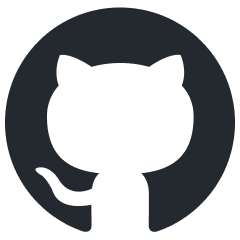}\hspace{2pt} \texttt{#1}}
\newcommand{\hflogo}[1]{\includegraphics[width=15pt, valign=c]{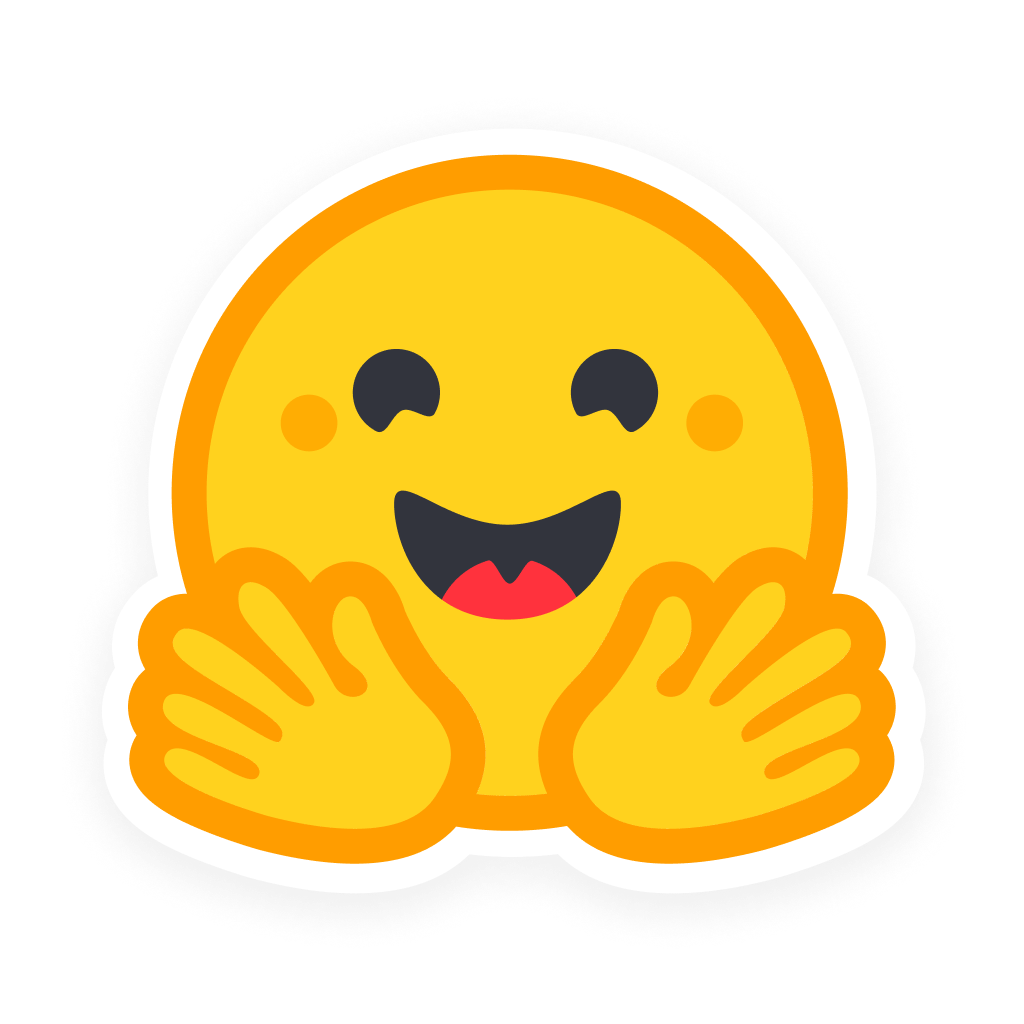} \texttt{#1}}

\usepackage[preprint]{icml2026}

\usepackage{amsmath}
\usepackage{amssymb}
\usepackage{mathtools}
\usepackage{amsthm}

\usepackage{multirow}
\usepackage{makecell}
\usepackage{enumitem}
\usepackage{tikz}
\usepackage[export]{adjustbox}
\newcommand{\tikzxmark}{%
	\tikz[scale=0.23] {
		\draw[line width=0.7,line cap=round] (0,0) to [bend left=6] (1,1);
		\draw[line width=0.7,line cap=round] (0.2,0.95) to [bend right=3] (0.8,0.05);
}}

\newcommand{\familyname}{Apertus-v1.1}
\newcommand{\apertustitle}{\resizebox{0.7\linewidth}{!}{\includegraphics{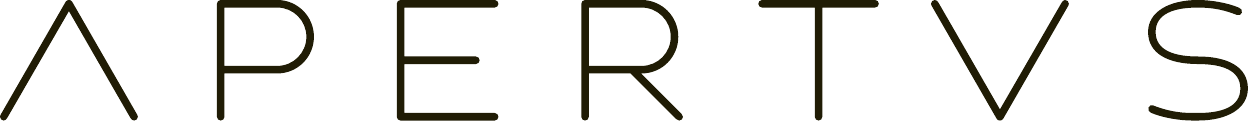}}}

\usepackage[capitalize,noabbrev]{cleveref}

\theoremstyle{plain}

\theoremstyle{definition}

\theoremstyle{remark}

\usepackage[textsize=tiny]{todonotes}

\icmltitlerunning{Apertus LLM Family Expansion via Distillation and Quantization}

\begin{document}

\twocolumn[
  \icmltitle{\apertustitle \\[1em] LLM Family Expansion via Distillation and Quantization}



  \icmlsetsymbol{equal}{*}

  \begin{icmlauthorlist}
    \icmlauthor{Andrei Panferov}{ISTA,EPFL}
    \icmlauthor{Davit Melikidze}{ETHZ}
    \icmlauthor{Martin Jaggi}{EPFL}
    \icmlauthor{Dan Alistarh}{ISTA,RedHatAI}
  \end{icmlauthorlist}

  \icmlaffiliation{ISTA}{ISTA}
  \icmlaffiliation{EPFL}{EPFL}
  \icmlaffiliation{ETHZ}{ETH Z\"{u}rich}
  \icmlaffiliation{RedHatAI}{Red Hat AI}

  \icmlcorrespondingauthor{Andrei Panferov}{andrei.panferov@ista.ac.at}

  \icmlkeywords{Machine Learning, ICML}

  \vskip 0.3in
]



\printAffiliationsAndNotice{}  

\begin{abstract}
The wide adoption of LLMs has led to their use in great variety of applications and scenarios, such as chatbot assistants and data annotation, creating the need for the models to satisfy certain budget and hardware constraints. This has led to the trend of LLMs being released in batches consisting of similar models of various sizes for the family of models to adhere to as wide of a range of constraints as possible. In this paper, we validate distillation and quantization as a cost-effective way to expand model families to new sizes and hardware formats. Based on the open-recipe Apertus 8B LLM, we produce \familyname~--- a distilled family of models with up to 4B parameters trained on 1.7T permissive license tokens. We demonstrate cost-efficiency and strong accuracy performance of our approach for covering large ranges of hardware and systems requirements.

\end{abstract}

\section{Background}

The popularity and versatility of Large Language Models (LLMs) have introduced a wide spectrum of budget, memory, and hardware constraints for their deployment. To accommodate these varying requirements, it has become crucial to provide LLMs in multiple sizes and formats. Releasing a family of models allows practitioners to select the optimal trade-off between computational cost and predictive performance for their specific deployment scenarios, democratizing access to advanced AI capabilities across different hardware tiers.

However, training an entire family of models from scratch requires prohibitive amounts of compute. Knowledge Distillation (KD) in the pre-training phase, or Pre-training Distillation (PD), offers a powerful solution to dramatically cut these costs \cite{peng2024pretrainingdistillationlargelanguage}. By transferring knowledge from a large, capable teacher model to a smaller student model using the teacher's generated logits, the student benefits from richer information and implicit label smoothing. This allows the student to converge faster and achieve higher downstream performance with significantly fewer training tokens and compute resources. Consequently, pre-training distillation enables the cost-effective expansion of a model family without the computational burden of standard pre-training.

An orthogonal direction for addressing cost requirements (e.g., disk space or latency) is quantization. While reducing numerical precision significantly lowers the memory footprint and accelerates inference, it inherently introduces a cost-accuracy trade-off. As we show here, by carefully balancing this trade-off around the Pareto frontier of compression methods, practitioners gain finer control over the model's performance and hardware profile. This fine-grained control allows for further expansion of the model family, bridging the gaps between pre-trained sizes at a cost significantly less than even pre-training distillation.

Our work builds upon the foundation of the Apertus~\citep{apertus2025apertusdemocratizingopencompliant} project, which sets a new standard for fully open and compliant LLMs. Unlike many open-weight models that withhold training data and pipelines, the Apertus recipe emphasizes complete transparency, data compliance, and global multilingual representation. By grounding our distillation and quantization pipeline in the Apertus ecosystem, we inherit its rigorous openness and reproducibility.


\begin{table*}[t]
\centering
\caption{Model architecture overview.}
\label{tab:architecture}
    \begin{tabular}{l|cccccc|rr}
    \toprule
    \multirow{2}{*}{Model} & \multirow{2}{*}{Layers} & \multirow{2}{*}{Dim} & \multirow{2}{*}{MLP Dim} & \multirow{2}{*}{Heads (Q/KV)} & \multirow{2}{*}{Dim/Layers} & \multirow{2}{*}{Tied Emb.} & \multicolumn{2}{c}{Model size} \\
    & & & & & & & Compute & Storage \\ \midrule
    \familyname-0.5B & 20     & 1024 & 6144    & 16/4          & 51.2           & Yes           & 0.4B                & 0.4B              \\
    \familyname-1.5B & 16     & 2048 & 12288   & 32/8          & 128            & No            & 1.5B                & 2.0B              \\
    \familyname-4B   & 24     & 3072 & 16384   & 24/8          & 128            & No            & 3.8B                & 4.6B              \\\hline
    Apertus-8B       & 32     & 4096 & 21504   & 32/8          & 128            & No            & 8.1B                & 9.1B              \\ \bottomrule
    \end{tabular}
\end{table*}


\section{Pre-Training Distillation}

\subsection{Recipe}

\paragraph{Data.} To produce the highest-quality models, we gathered the data corresponding to Phase 5 (the final phase) of the original Apertus pre-training, which consists of documents and code and instruction samples with the highest level of quality filtering for a total yield of approximately 1.7T tokens. Similar to Apertus, we cut and pack these documents into chunks of 4096 tokens and train with cross-document attention masked.

\paragraph{Logits generation.} To be able to efficiently re-use the logits for multiple models, we generated the entire training set in advance. We ran the collected documents through the \texttt{Apertus-8B-2509} model to obtain $\approx$131k logits per token. After calculating the probability distributions from these logits, top-256 highest probabilities were identified per token. These probabilities, along with corresponding token indices in model vocabulary, were represented in 32-bit precision for a total of $\approx$2KB of data per token. The tensors were batched in groups of $\approx$131k tokens, compressed with \texttt{gzip} and stored in long-term storage for a total footprint of $\approx$1.5PB of disk space. We applied sequences permutation on the logits generation stage to only have to do sequential disk loads when using them for training later.

\paragraph{Training objective.} As shown to perform well by~\citet{peng2024pretrainingdistillationlargelanguage}, we utilize a 90\%/10\% mix between the KL-Divergence and the label cross-entropy. Since the computed KL-Divergence is sparse, it introduces close to no computational or memory overhead relative to the basic cross-entropy calculation.

\paragraph{Model Architecture.}

\familyname~models follow the same architecture as Apertus: Dense transformer models with grouped-query attention and xIELU~\citep{huang2025derivingactivationfunctionsusing} activation in the MLP. Table~\ref{tab:architecture} details the architectural configurations, parameter counts, and the resulting memory and computational footprints for the \familyname~models. Notably, we used tied embeddings and thinner and deeper architecture for the smallest \familyname~model to maximize performance while minimizing memory footprint~\citep{liu2024mobilellmoptimizingsubbillionparameter}. 

\begin{figure}[h]
    \centering
    \includegraphics[width=1.0\linewidth]{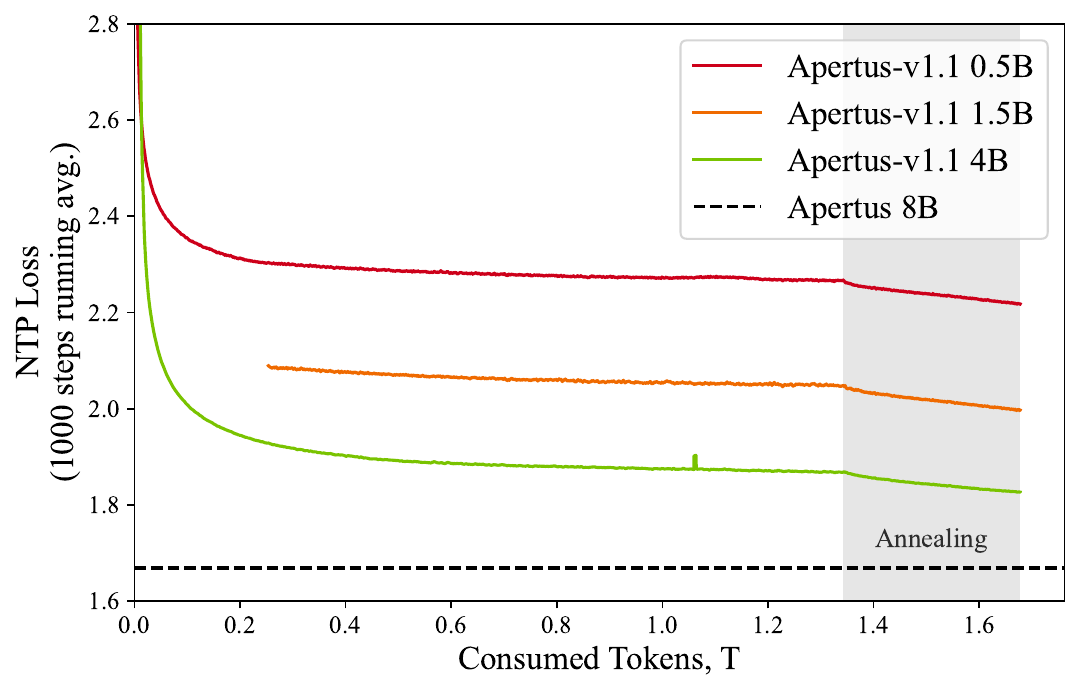}
    \caption{Training loss curves of \familyname~models. Dashed line shows the loss of the teacher model (\texttt{Apertus-8B-2509}).}
    \label{fig:pretrain}
\end{figure}

\begin{figure}[h]
    \centering
    \includegraphics[width=1.0\linewidth]{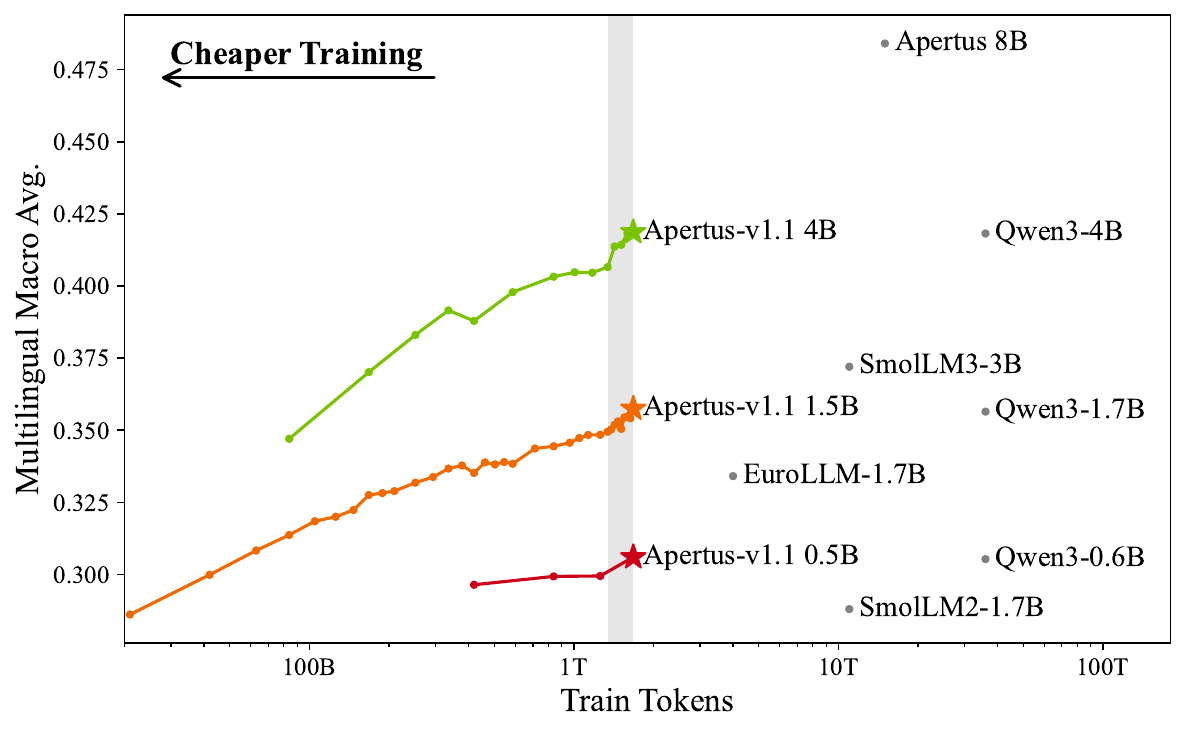}
    \caption{Multilingual performance macro average during pre-training of \familyname~models and for a number of similar-sized models. Distillation allows \familyname~models to achieve competitive performance while training on up to an order of magnitude less compute.}
    \label{fig:pretrain_metrics}
\end{figure}

\paragraph{Training dynamics.} Similar to Apertus, we use the AdEMAMix~\citep{ICLR2025_a2cf225b} optimizer with WSD schedule and weight decay. Next-token prediction (NTP) loss shown in Figure~\ref{fig:pretrain}. Multilingual macro downstream average shown in Figure~\ref{fig:pretrain_metrics}. We observed no training instabilities and consistent improvement in downstream performance, especially during the learning rate annealing stage (highlighted in gray).

\paragraph{SFT and alignment.} The supervised fine-tuning (SFT) stage followed immediately after pre-training. For it, we exactly reused the original Apertus SFT recipe, only adjusting the LR to match the post-annealing LR of \familyname~models. For the subsequent alignment stage, we utilized a simplified DPO~\citep{rafailov2024directpreferenceoptimizationlanguage} setup.

\paragraph{Evaluations.} Following the Apertus evaluation setup, we report multilingual benchmarks average during training in Figure~\ref{fig:pretrain_metrics}, selected final pre-training metrics in Table~\ref{tab:pretrain_benchmarks}, multilingual post-training evaluations in Table~\ref{tab:instruct_benchmarks} and broader post-training evaluations in Appendix~\ref{app:evals}. Unsurprisingly, the performance profile of \familyname~models is extremely similar to \texttt{Apertus-8B-2509}, demonstrating great multilingual performance for the base models and good multilingual chat performance but lacking in certain capabilities like instruction following and math.

\subsection{Cost Analysis}

\begin{table}[h!]
    \centering
    \caption{Cost for small LLM pre-training and distillation. \familyname~is 2-10x cheaper than competing small LLM pre-training pipelines.}
    \label{tab:cost}
    \small
    \begin{tabular}{l|cc}
        \toprule
        \textbf{Stage} & \textbf{Tokens} & \textbf{FLOPs} \\
        \midrule
        Original pre-training  & \multirow{2}{*}{15T} & \multirow{2}{*}{3.7E23} \\
        Apertus-8B    &  &  \\\midrule\midrule
        Logits generation  & \multirow{2}{*}{1.7T} & \multirow{2}{*}{1.4E22} \\
        from Apertus-8B    &  &  \\\midrule
        Pre-training  & \multirow{2}{*}{1.7T} & \multirow{2}{*}{0.2E22} \\
        \familyname~0.5B    &  &  \\\hline
        Pre-training  & \multirow{2}{*}{1.7T} & \multirow{2}{*}{0.8E22} \\
        \familyname~1.5B    &  &  \\\hline
        Pre-training  & \multirow{2}{*}{1.7T} & \multirow{2}{*}{2.0E22} \\
        \familyname~4B    &  &  \\\midrule\midrule
        Pre-training  & \multirow{2}{*}{36T} & \multirow{2}{*}{6.5E22} \\
        Qwen3-0.6B    &  &  \\\hline
        Pre-training  & \multirow{2}{*}{4T} & \multirow{2}{*}{1.7E22} \\
        EuroLLM-1.7B    &  &  \\\hline
        Pre-training  & \multirow{2}{*}{11T} & \multirow{2}{*}{5.6E22} \\
        SmolLM2-1.7B    &  &  \\\hline
        Pre-training  & \multirow{2}{*}{11T} & \multirow{2}{*}{9.9E22} \\
        SmolLM3-3B    &  &  \\
        \bottomrule
    \end{tabular}
\end{table}

As seen from Table~\ref{tab:cost}, \familyname~models used significantly less compute than similar-sized models, being trained on just 1.7T tokens, in contrast to the 15T tokens of Apertus. The cost of producing the logits from the 8B model is relatively small because one only needs to perform the forward pass to produce logits and the same logits only have to be computed once for the entire family of distilled models, dramatically cutting the teacher cost per-model. The total compute cost of the entire \familyname~model family is 2.4E22 FLOPs. This is less than, for example,  the cost of standalone  pre-training for SmolLM2-1.7B and less than 12\% of the original Apertus 8B pre-training cost.

\begin{table*}[ht]
    \centering
    \caption{Base models evaluations.}
    \label{tab:pretrain_benchmarks}
    \begin{tabular}{lc|cccccc}
        \toprule
        \textbf{Model} & \textbf{Avg} & \textbf{ARC} & \textbf{HellaSwag} & \textbf{WinoGrande} & \textbf{XNLI} & \textbf{XCOPA} & \textbf{PIQA} \\
        \midrule
        \familyname-0.5B  & 51.79 & 44.96 & 40.42 & 57.06 & 41.51 & 55.49 & 71.27 \\
        \familyname-1.5B  & 56.66 & 52.66 & 48.31 & 61.72 & 42.94 & 59.76 & 74.54 \\
        \familyname-4B    & 61.53 & 61.15 & 53.51 & 67.48 & 45.03 & 63.82 & 78.18 \\\hline
        Apertus-8B    & 64.96 & 71.66 & 59.62 & 69.30 & 44.09 & 65.69 & 79.38 \\
        \midrule
        EuroLLM-1.7B    & 54.03 & 50.80 & 45.01 & 59.51 & 40.88 & 55.76 & 72.20 \\
        SmolLM2-1.7B    & 58.00 & 60.23 & 53.38 & 66.22 & 37.57 & 53.51 & 77.10 \\
        SmolLM-3B-Base  & 60.88 & 64.45 & 56.37 & 68.43 & 40.28 & 58.02 & 77.75 \\
        Qwen3-0.6B-Base & 52.23 & 48.35 & 41.01 & 59.20 & 39.55 & 54.96 & 70.29 \\
        Qwen3-1.7B-Base & 57.51 & 56.49 & 49.36 & 63.38 & 41.66 & 58.35 & 75.79 \\
        Qwen3-4B-Base   & 62.14 & 64.99 & 54.56 & 70.48 & 43.00 & 61.82 & 77.97 \\
        \bottomrule
    \end{tabular}
\end{table*}

\begin{table*}[ht]
    \centering
    \caption{Multilingual evaluations for instruction-tuned models. Each benchmark here is the multilingual version thereof (see Appendix~\ref{app:evals}).}
    \label{tab:instruct_benchmarks}
    \begin{tabular}{lc|ccccc}
        \toprule
        \textbf{Model} & \textbf{Average} & \textbf{MMLU} & \textbf{TruthfulQA} & \textbf{Arc} & \textbf{IF} & \textbf{LogiQA} \\
        \midrule
        \familyname-0.5B Instruct & 0.318 & 0.258 & 0.461 & 0.225 & 0.328 & 0.279 \\
        \familyname-1.5B-Instruct & 0.382 & 0.377 & 0.451 & 0.266 & 0.434 & 0.276 \\
        \familyname-4B-Instruct   & 0.473 & 0.504 & 0.506 & 0.332 & 0.550 & 0.296 \\\hline
        Apertus-8B-Instruct-2509 & 0.534 & 0.553 & 0.524 & 0.368 & 0.689 & 0.290 \\
        \midrule
        EuroLLM-1.7B-Instruct & 0.291 & 0.260 & 0.433 & 0.250 & 0.222 & 0.269 \\
        EuroLLM-9B-Instruct	  & 0.480 & 0.520 & 0.465 & 0.322 & 0.613 & 0.345 \\
        gemma-3-270m-it       & 0.289 & 0.242 & 0.465 & 0.215 & 0.236 & 0.205 \\
        gemma-3-1b-it         & 0.406 & 0.409 & 0.457 & 0.250 & 0.509 & 0.379 \\
        gemma-3-4b-it         & 0.497 & 0.547 & 0.492 & 0.316 & 0.635 & 0.411 \\
        SmolLM2-1.7B-Instruct & 0.348 & 0.365 & 0.452 & 0.213 & 0.364 & 0.246 \\
        SmolLM3-3B            & 0.479 & 0.507 & 0.500 & 0.270 & 0.637 & 0.365 \\
        Qwen3-0.6B            & 0.401 & 0.377 & 0.464 & 0.222 & 0.541 & 0.353 \\
        Qwen3-1.7B            & 0.457 & 0.477 & 0.490 & 0.251 & 0.611 & 0.414 \\
        Qwen3-4B              & 0.521 & 0.581 & 0.497 & 0.274 & 0.733 & 0.500 \\
        \bottomrule
    \end{tabular}
\end{table*}

\section{Quantization}

While pre-training distillation successfully generated the core \familyname models at a fraction of the cost, adapting these models for highly constrained environments requires further optimization for specific hardware profiles. In this section, we consider the problem of adapting \familyname~models to NVIDIA GPUs and mobile devices, demonstrating how quantization yields a wider range of optimal, specialized models at close to no cost.

\begin{figure*}[t]
    \centering
    \includegraphics[width=1.0\linewidth]{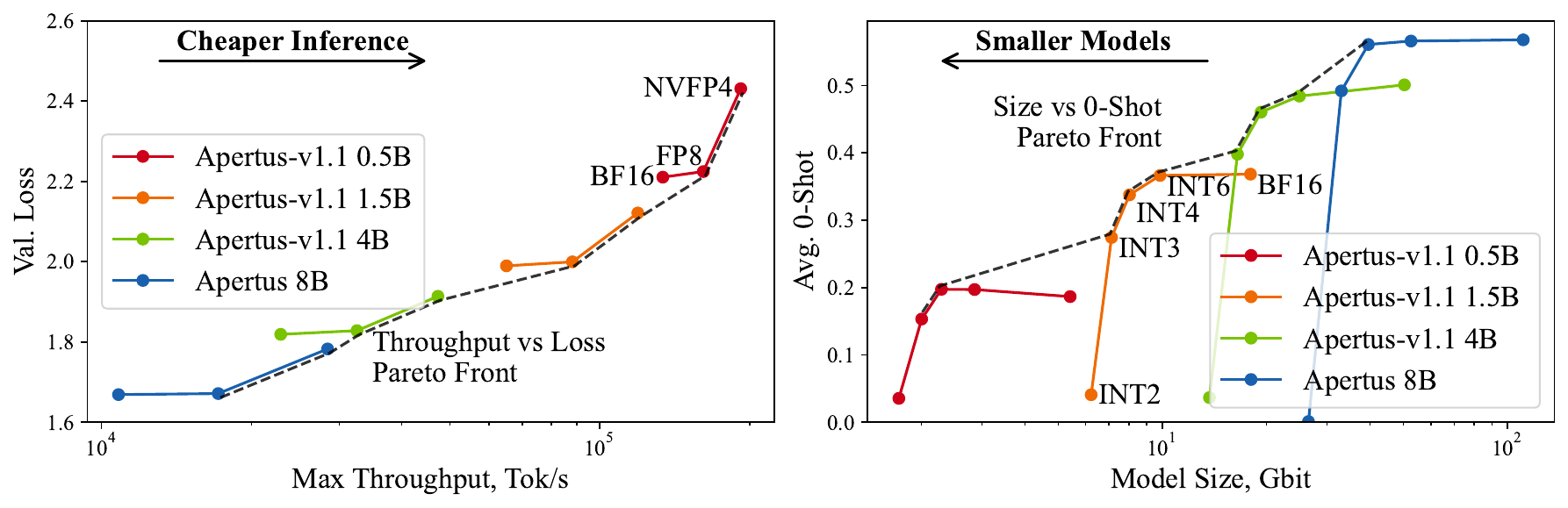}
    \caption{Visualization of the cost-accuracy trade-off for Apertus and \familyname~models. Base models (left) are compared based on validation loss while instruction-tuned models (right) are compared based on downstream performance. Quantized models both optimize the trade-off and add intermediate points to the Pareto fronts.}
    \label{fig:pareto}
\end{figure*}

\subsection{\familyname~Quantization Recipe}

\begin{figure}
    \centering
    \includegraphics[width=1.0\linewidth]{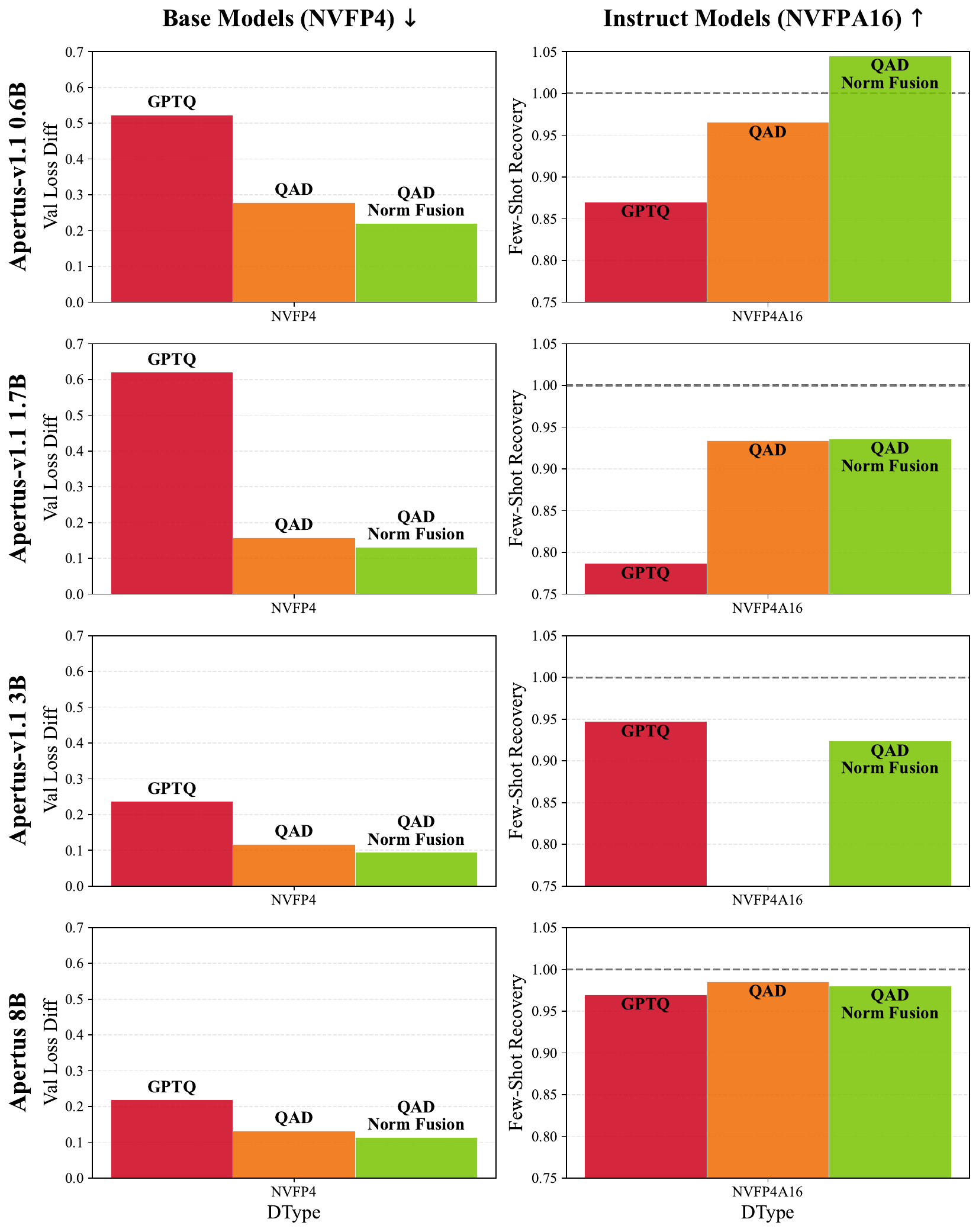}
    \caption{\familyname~quantization recipe ablation.}
    \label{fig:combined_normalization_results}
\end{figure}

\paragraph{Baseline.} We use GPTQ~\citep{frantar2023gptqaccurateposttrainingquantization}, the most widely-used 1-shot LLM quantization method as our baseline. We gauge our improvement over it differently for base and instruction-tuned models:

\begin{itemize}
    \item \textbf{For base models,} we measure the loss increase over the corresponding unquantized models on the validation set of $\approx$17M tokens from the original pre-training mixture (Apertus Phase 5 data). We test \emph{weight+activation} (FP8, NVFP4) quantization for base models with focus on NVIDIA Blackwell GPUs, as we foresee their main usage in high-throughput scenarios such as data annotation and embedding.
    \item \textbf{For instruction-tuned models,} we measure the recovery of macro average over normalized few-shot accuracies on Arc~\citep{clark2018thinksolvedquestionanswering}, HellaSwag~\citep{zellers2019hellaswagmachinereallyfinish}, MMLU~\citep{hendrycks2021measuringmassivemultitasklanguage} and WinoGrande~\citep{sakaguchi2019winograndeadversarialwinogradschema}. We test \emph{weight-only quantization} (INT2, INT3, INT4, INT6) for instruction-tuned models with focus on Apple devices (MLX) inference, as we foresee their main usage in memory-limited scenarios such as mobile and edge deployment.
\end{itemize}

\paragraph{Quantization-aware distillation (QAD).} QAD is applied as a short recovery stage on a fully-trained model by treating the entire model as trainable parameters, quantizing its weights every forward pass and updating them with standard gradient-based method via straight-through estimation~\citep{bengio2013estimatingpropagatinggradientsstochastic}, bridging the gap between full quantization-aware training and PTQ methods. Similar to pre-training distillation, teacher model logits (usually the corresponding unquantized model or a larger model from the same family) provide much richer signal for this phase, making it preferable to quantization-aware supervised fine-tuning. QAD has been shown to yield consistent improvement over 0-shot and 1-shot post-training quantization (PTQ) methods~\citep{lee2025unifyingblockwiseptqdistillationbased,egiazarian2026bridginggappromiseperformance,xin2026quantizationawaredistillationnvfp4inference}.

The open access to the original pre-training set and SFT mixture utilized for both Apertus and \familyname~pre- and post-training allows us to use it for QAD of these models with the highest degree of confidence that the distillation curriculum captures close to the entirety of the models' capability. We test QAD for both base and instruction-tuned models, using $\approx$100M tokens (we see only marginal improvement beyond that) of the pre-training or the SFT mixture accordingly. We use \texttt{Apertus-8B-2509} and \texttt{Apertus-8B-Instruct-2509} as a teacher in this scenario. Additional implementation details and hyper-parameters are described in Appendix~\ref{app:qad}.

\paragraph{Norm fusion.} To further improve quantization quality, we propose the following zero-cost static model optimization: We scale attention's QKV and MLP's up projection matrices' columns (input dimension) to have the same norm, multiplicatively fusing the reciprocal scales into the preceding layer-normalization layers' weights. The idea behind this is to normalize the magnitudes of weight values to prevent flush-to-zero of small-magnitude but important weights and weights adjacent to outlier channels.

The loss measurements for compressed base models and few-shot recovery measurements for the instruction-tuned models show that this yields the most improvement for smaller models. Additionally, although this normalization is mainly designed to assist with weight quantization, we find that it also improves weight+activation quantization (NVFP4), indicating that offloading these scales to activations doesn't hurt their compressibility.

\begin{figure*}[t]
    \centering
    \includegraphics[width=1.0\linewidth]{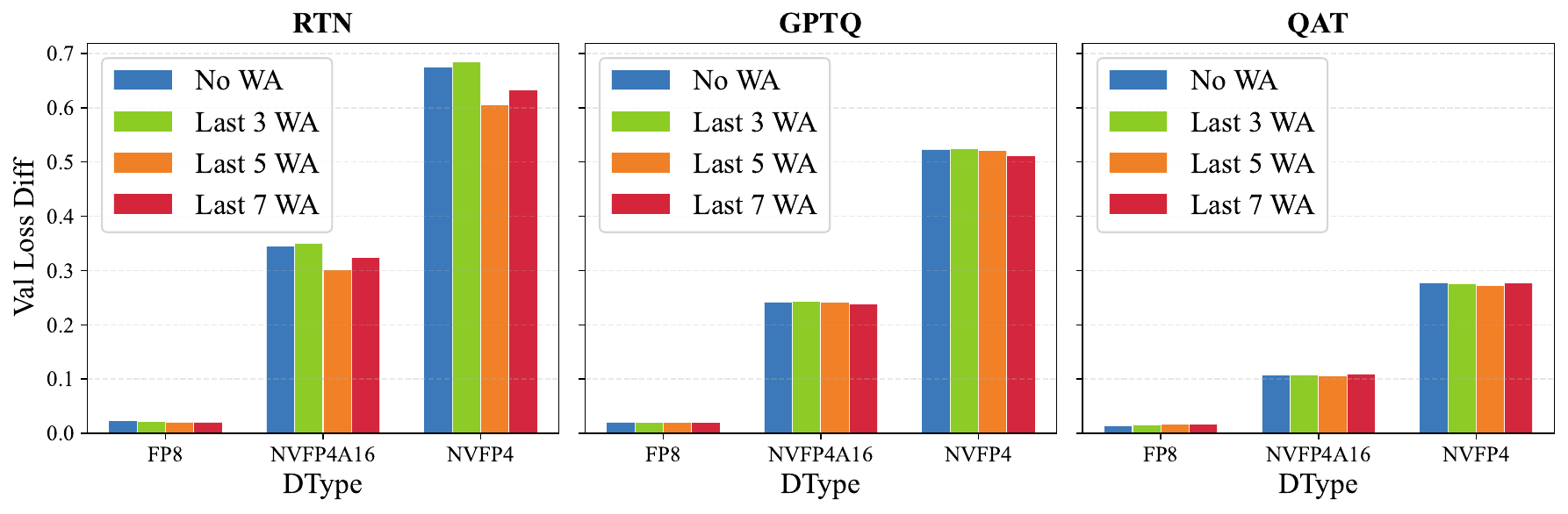}
    \caption{The effect of weight averaging (WA) over the last few base model checkpoints on post-training quantization for various data-types and algorithms. Checkpoints were taken every 1000 iterations.}
    \label{fig:0.6B_avg_comparison}
\end{figure*}

\paragraph{Weight averaging.}
Weight averaging (arithmetic averaging of model weight tensors) of the last few checkpoints during the annealing stage has been shown to improve LLMs' resilience to post-training quantization~\citep{ajroldi2025whenaverageweights}.  To validate it, we tested weight averaging for the \familyname~0.5B base model combined with various quantization formats and methods, including RTN, GPTQ~\citep{frantar2023gptqaccurateposttrainingquantization} and QAD. The results, shown in Figure~\ref{fig:0.6B_avg_comparison}, demonstrate that weight averaging reduces validation loss gap to BF16 by up to 10\% for RTN, up to 2\% for GPTQ and has close to \emph{no discernible effect on QAD}. As a result, we did not include it in our final quantization pipeline. 

\paragraph{Final quantization recipe.} Our final recipe combines QAD with norm fusion to achieve just 0.1-0.2 validation loss increase for base and 90-104\% few-shot accuracy recovery for instruction-tuned Apertus and \familyname~models, as seen in Figure~\ref{fig:combined_normalization_results}.

\subsection{Pareto Optimality}

As mentioned in the beginning of this section, we analyze base model quantization in the context of high-throughput applications and instruction-tuned model quantization in the context of memory-constrained deployment. Naturally, the corresponding cost can be measured for every model we trained (quantized or otherwise), along with a representative measure of it's capability, quantifying the cost-accuracy trade-off. Covering a larger range of costs is what drove the demand for smaller models in the first place, and in Figure~\ref{fig:pareto} one can see this trade-off visualized.

What is interesting, is that quantized models not only shift the Pareto front (i.e., the enveloping curve) towards more efficient solutions (as seen, for example, by BF16 models almost never being optimal), but also adds more points on the frontier, allowing for more fine-grained control over cost. Without quantization, adding new points would have meant pre-training new models of intermediate sizes, which would have entailed spending compute in the order of trillions tokens. QAD, on the other hand, achieves high recovery after only a few tens of millions of tokens, cutting the cost by more than \emph{four orders of magnitude}.

\section{Released Checkpoints}
\label{sec:released_checkpoints}

We provide a comprehensive suite of pre-trained and instruction-tuned models across multiple quantization formats to support various hardware constraints and deployment scenarios. Table~\ref{tab:released_checkpoints} summarizes all the checkpoints released as part of the Apertus and \familyname~model families. 

\begin{table*}[ht]
    \centering
    \caption{Overview of released Apertus and \familyname~checkpoints. Click the Hugging Face logo to access the corresponding model weights.}
    \label{tab:released_checkpoints}
    \begin{tabular}{l|c|cccccc}
        \toprule
        \multirow{2}{*}{\textbf{Model}} & \textbf{BF16} & \textbf{BF16} & \textbf{FP8} & \textbf{NVFP4A16} & \textbf{INT3} & \textbf{INT4} & \textbf{INT6} \\
         & Base & \multicolumn{6}{c}{Instruct}  \\
        \midrule
        \textbf{\familyname-0.5B} & \href{https://huggingface.co/swiss-ai/Apertus-v1.1-0.5B}{\hflogo{}} & \href{https://huggingface.co/swiss-ai/Apertus-v1.1-0.5B-Instruct}{\hflogo{}} & \href{https://huggingface.co/swiss-ai/Apertus-v1.1-0.5B-Instruct-vLLM-FP8}{\hflogo{}} & \href{https://huggingface.co/swiss-ai/Apertus-v1.1-0.5B-Instruct-vLLM-NVFP4A16}{\hflogo{}} & \href{https://huggingface.co/swiss-ai/Apertus-v1.1-0.5B-Instruct-MLX-INT3}{\hflogo{}} & \href{https://huggingface.co/swiss-ai/Apertus-v1.1-0.5B-Instruct-MLX-INT4}{\hflogo{}} & \href{https://huggingface.co/swiss-ai/Apertus-v1.1-0.5B-Instruct-MLX-INT6}{\hflogo{}} \\
        \textbf{\familyname-1.5B} & \href{https://huggingface.co/swiss-ai/Apertus-v1.1-1.5B}{\hflogo{}} & \href{https://huggingface.co/swiss-ai/Apertus-v1.1-1.5B-Instruct}{\hflogo{}} & \href{https://huggingface.co/swiss-ai/Apertus-v1.1-1.5B-Instruct-vLLM-FP8}{\hflogo{}} & \href{https://huggingface.co/swiss-ai/Apertus-v1.1-1.5B-Instruct-vLLM-NVFP4A16}{\hflogo{}} & \href{https://huggingface.co/swiss-ai/Apertus-v1.1-1.5B-Instruct-MLX-INT3}{\hflogo{}} & \href{https://huggingface.co/swiss-ai/Apertus-v1.1-1.5B-Instruct-MLX-INT4}{\hflogo{}} & \href{https://huggingface.co/swiss-ai/Apertus-v1.1-1.5B-Instruct-MLX-INT6}{\hflogo{}} \\
        \textbf{\familyname-4B} & \href{https://huggingface.co/swiss-ai/Apertus-v1.1-4B}{\hflogo{}} & \href{https://huggingface.co/swiss-ai/Apertus-v1.1-4B-Instruct}{\hflogo{}} & \href{https://huggingface.co/swiss-ai/Apertus-v1.1-4B-Instruct-vLLM-FP8}{\hflogo{}} & \href{https://huggingface.co/swiss-ai/Apertus-v1.1-4B-Instruct-vLLM-NVFP4A16}{\hflogo{}} & \href{https://huggingface.co/swiss-ai/Apertus-v1.1-4B-Instruct-MLX-INT3}{\hflogo{}} & \href{https://huggingface.co/swiss-ai/Apertus-v1.1-4B-Instruct-MLX-INT4}{\hflogo{}} & \href{https://huggingface.co/swiss-ai/Apertus-v1.1-4B-Instruct-MLX-INT6}{\hflogo{}} \\
        \textbf{Apertus-8B-2509} & \href{https://huggingface.co/swiss-ai/Apertus-8B-2509}{\hflogo{}} & \href{https://huggingface.co/swiss-ai/Apertus-8B-Instruct-2509}{\hflogo{}} & \href{https://huggingface.co/swiss-ai/Apertus-8B-Instruct-2509-vLLM-FP8}{\hflogo{}} & \href{https://huggingface.co/swiss-ai/Apertus-8B-Instruct-2509-vLLM-NVFP4A16}{\hflogo{}} & \tikzxmark & \href{https://huggingface.co/swiss-ai/Apertus-8B-Instruct-2509-MLX-INT4}{\hflogo{}} & \tikzxmark \\
        \bottomrule
    \end{tabular}
\end{table*}

\section{Conclusion}

We validated pre-training distillation for multi-billion parameter models and multi-trillion token budgets, demonstrating how such model family expansion can be done at a tiny cost (less than 20\%) of the teacher model training and far more cheaply than pre-training from scratch. In total, we release 24 new model checkpoints, including the 3 pre-trained base models, 3 instruction-tuned models, 8 quantized checkpoints for NVIDIA devices, 10 quantized checkpoints for Apple devices, as well as all the code to reproduce training, post-training and quantization pipelines.

We hope our open-source, open-data and compliant recipe to be of use for LLM practitioners interested in producing and using small language models.

\section*{Acknowledgments}

We would like to thank Dhia Garbaya for his help with configuring the models and Hanna Yukhymenko for tuning the post-training setup.
This work was supported under project IDs a140 and infra01 as part of the Swiss AI Initiative, through a grant from the ETH Domain and computational resources provided by the Swiss National Supercomputing Centre (CSCS) under the Alps infrastructure. This research was funded in whole or in part by the Austrian Science Fund (FWF) 10.55776/COE12.


\bibliography{example_paper}
\bibliographystyle{icml2026}

\newpage
\appendix
\onecolumn
\section{Codebases}
\label{app:codebases}

The full codebases for the pre-training distillation, post-training, evaluations and quantization stages of the pipeline are available on GitHub.

\begin{itemize}
    \item \href{https://github.com/swiss-ai/Megatron-LM-Distill}{\githublogo{} \texttt{Megatron-LM-Distill}}: A fork of Megatron-LM with added functionality for teacher logits generation and saving as well as pre-training distillation.
    \item \href{https://github.com/swiss-ai/posttraining}{\githublogo{} \texttt{posttraining}}: The original post-training codebase from Apertus that was reused for this project.
    \item \href{https://github.com/swiss-ai/qat-suite}{\githublogo{} \texttt{qat-suite}}: A lightweight quantization suite with support for vLLM and MLX data formats and various quantization algorithms, including QAD.
    \item \href{https://github.com/swiss-ai/evals}{\githublogo{} \texttt{evals}}: The Apertus pre-training evaluation suite.
    \item \href{https://github.com/swiss-ai/evals-post-train}{\githublogo{} \texttt{evals-post-train}}: The Apertus post-training evaluation suite.
\end{itemize}

For the data preparation scripts, please refer to the original Apertus report~\citep{apertus2025apertusdemocratizingopencompliant}.

\section{Evaluation Suite Details}
\label{app:evals}

For the evaluations reported in Tables~\ref{tab:pretrain_benchmarks} and~\ref{tab:instruct_benchmarks}, we used the publicly-available Apertus evaluation suite. The multilingual macro average shown in Figure~\ref{fig:pretrain_metrics} includes INCLUDE~\citep{romanou2025include}, XCOPA~\citep{ponti2020xcopamultilingualdatasetcausal}, XNLI~\citep{conneau2018xnli}, XWinograd~\citep{muennighoff2022crosslingual}, PAWS-X~\citep{yang-etal-2019-paws}, Multilingual Arc~\citep{dac2023okapi}, Global MMLU~\citep{singh2025globalmmluunderstandingaddressing} and Multilingual HellaSwag~\citep{dac2023okapi}.

\section{Additional Hyper-Parameters}

\subsection{Pre-Training Details}
\label{app:pre-training}

Additional per-model pre-training hyper-parameters are shown in Table~\ref{tab:exta_hp}.

\begin{table*}[t]
\centering
\caption{Additional hyper-parameters.}
\label{tab:exta_hp}
    \begin{tabular}{l|cccc}
    \toprule
    Model & LR & GBS & Total Iterations \\ \midrule
    \familyname-0.5B & 6e-4     & 512 & 800000 \\
    \familyname-1.5B & 3e-4     & 512 & 800000 \\
    \familyname-4B   & 2e-4     & 1024 & 400000 \\\bottomrule
    \end{tabular}
\end{table*}

\subsection{QAT Details}
\label{app:qad}

For the base models, we sample $\approx$130M tokens uniformly from the unused remainder of the gathered pre-training data. For the instruction-tuned models, we sample $\approx$60M uniformly from the Apertus SFT mixture. We train with AdamW~\citep{loshchilov2019decoupledweightdecayregularization} with cosine LR schedule. For base models, we use the same sequence length and batch size as in pre-training. For instruction-tuned models, we use slightly larger batch size of 512-2048 to compensate for smaller length of some post-training sequences. Similar to pre-training distillation, we pre-compute and store the sparse logits from the teacher model (\texttt{Apertus-8B-2509} for base models and \texttt{Apertus-8B-Instruct-2509} for instruction-tuned models) once and re-use them for all student model and quantization format combinations.


\end{document}